\documentclass{article}

\usepackage{arxiv}

\usepackage[utf8]{inputenc}
\usepackage[T1]{fontenc}   
\usepackage{hyperref}      
\usepackage{url}            
\usepackage{booktabs}       
\usepackage{amsfonts}       
\usepackage{nicefrac}       
\usepackage{microtype}      
\usepackage{graphicx}
\usepackage[numbers]{natbib}
\usepackage{doi}
\usepackage{amsmath}
\usepackage{multirow}
\usepackage{cleveref}
\usepackage{todonotes}
\usepackage{subcaption}
\usepackage{lscape} 
\usepackage{lipsum}
\usepackage{algorithm}
\usepackage{algorithmic}
\usepackage{float}

\renewcommand{\headeright}{}
\renewcommand{\undertitle}{}

\title{SPIKANs: Separable Physics-Informed Kolmogorov-Arnold Networks}

\author{
        \href{https://orcid.org/0009-0001-5361-3105}{\includegraphics[scale=0.06]{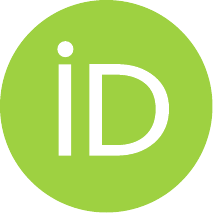}\hspace{1mm}Bruno Jacob}\\
	Pacific Northwest National Laboratory\\
	Richland, WA 99354 \\
	\texttt{bruno.jacob@pnnl.gov} \\
      \And  
        \href{https://orcid.org/0000-0002-6411-6198}{\includegraphics[scale=0.06]{Figures/orcid.pdf}\hspace{1mm}Amanda A. Howard}\\
	Pacific Northwest National Laboratory\\
	Richland, WA 99354 \\
	\texttt{amanda.howard@pnnl.gov} \\
     \And  
        \href{https://orcid.org/0000-0002-9928-5637}{\includegraphics[scale=0.06]{Figures/orcid.pdf}\hspace{1mm}Panos Stinis}\\
	Pacific Northwest National Laboratory\\
	Richland, WA 99354 \\
	\texttt{panagiotis.stinis@pnnl.gov} \\
}

\begin{document}

\maketitle

\begin{abstract}
Physics-Informed Neural Networks (PINNs) have emerged as a promising method for solving partial differential equations (PDEs) in scientific computing. While PINNs typically use multilayer perceptrons (MLPs) as their underlying architecture, recent advancements have explored alternative neural network structures. One such innovation is the Kolmogorov-Arnold Network (KAN), which has demonstrated benefits over traditional MLPs, including faster neural scaling and better interpretability. The application of KANs to physics-informed learning has led to the development of Physics-Informed KANs (PIKANs), enabling the use of KANs to solve PDEs. However, despite their advantages, KANs often suffer from slower training speeds, particularly in higher-dimensional problems where the number of collocation points grows exponentially with the dimensionality of the system. To address this challenge, we introduce Separable Physics-Informed Kolmogorov-Arnold Networks (SPIKANs). This novel architecture applies the principle of separation of variables to PIKANs, decomposing the problem such that each dimension is handled by an individual KAN. This approach drastically reduces the computational complexity of training without sacrificing accuracy, facilitating their application to higher-dimensional PDEs. Through a series of benchmark problems, we demonstrate the effectiveness of SPIKANs, showcasing their superior scalability and performance compared to  PIKANs and highlighting their potential for solving complex, high-dimensional PDEs in scientific computing.
\end{abstract}


\section{Introduction}
Physics-Informed Neural Networks (PINNs) have gained widespread attention for their ability to incorporate physical laws into machine learning models, allowing solutions to forward and inverse problems involving partial differential equations (PDEs) \cite{raissi2019physics}. However, traditional PINNs struggle with computational costs when solving multi-dimensional, highly complex PDEs. This challenge is exacerbated by the need for a large number of collocation points, leading to high memory overhead and inefficient scaling as the problem’s dimensionality increases \cite{karniadakis2021physics}.

As an alternative to the traditional multi-layer perceptron (MLP) architecture, Kolmogorov-Arnold networks (KANs) introduced by \cite{liu2024kan} and extended in \cite{liu2024kan20} have prompted further exploration in the context of physics-informed learning. These networks, featuring trainable activation functions, have demonstrated promising results and superior performance compared to MLPs in terms of interpretability, robustness against catastrophic forgetting \cite{vaca2024kolmogorov, samadi2024smooth}, and resilience to noisy data \cite{howard2024finite}. To leverage these benefits within the PINNs framework, \cite{liu2024kan, shukla2024comprehensive} have proposed physics-informed KANs (PIKANs) and deep operator networks (DeepOKANs \cite{abueidda2024deepokan}). PIKANs have demonstrated success on benchmark problems for physics-informed machine learning and applications \cite{rigas2024adaptive, ranasinghe2024ginn,wang2024kolmogorov, toscano2024inferring, patra2024physics, shuai2024physics, so2024higher}.

Despite the advantages of KANs in physics-informed learning, the use of B-splines as basis functions for the activations in the original formulation leads to a significant increase in computational cost. Although previous works have proposed alternative basis functions \cite{bozorgasl2024wav, li2024kolmogorov, ss2024chebyshev, yu2024sinc} that reduce computational time per iteration, the fundamental issue of exponential growth in the number of training points and network evaluations remains. Consequently, the curse of dimensionality associated with this architecture, combined with slower performance in KANs, limits the application of these networks for solving high-dimensional initial-boundary value problems.

To address these limitations, we present Separable Physics-Informed Kolmogorov-Arnold Networks (SPIKANs). Inspired by Separable PINNs \cite{cho2024separable}, we propose a network architecture that decomposes the solution of multi-dimensional PDEs into separable components, enabling more efficient training and inference. For a $d$-dimensional PDE with $N^d$ sampled collocation points, instead of training one KAN with a cloud of $O(N^d)$ training points, SPIKANs decompose the problem into $d$ KANs, each receiving $O(N)$ points as inputs. This approach substantially reduces computational costs in terms of time and memory, at the cost of requiring a factorizable mesh of collocation points rather than the traditional unstructured point-cloud used in PINNs.

This paper is organized as follows: in Sec.~\ref{sec:methods}, we briefly introduce KANs and  PIKANs, along with the description of the proposed method, SPIKANs. In Sec.~\ref{sec:results}, we demonstrate the use of SPIKANs in four benchmark problems, in increasing order of dimensionality, comparing gains in accuracy and speedup. Finally, we summarize the findings and discuss the limitations in Sec.~\ref{sec:conclusions}. 

\section{Methods}
\label{sec:methods}

\subsection{Kolmogorov-Arnold Networks (KANs)}

Kolmogorov-Arnold Networks (KANs) are a type of neural architecture proposed by \cite{liu2024kan}. They are inspired by the Kolmogorov-Arnold representation theorem, which states that a multivariate continuous function $u(\boldsymbol{x}) = u(x_1, x_2, \dots, x_n)$ defined on a bounded domain can be expressed as a composition of continuous univariate functions and sums \cite{shukla2024comprehensive}. In practice, \cite{liu2024kan} proposes approximating $u$ as

\begin{equation}
u(x_1, \ldots, x_n) = \sum_{i_{L-1}=1}^{n_{L-1}} \phi_{L-1, i_L, i_{L-1}} \left( \sum_{i_{L-2}=1}^{n_{L-2}} \dots \left(\sum_{i_0 = 1}^{n_0} \phi_{0,i_1, i_0}(x_{i_0}) \right) \right) \label{eq:kan_approximation},
\end{equation}

where $L$ denotes the number of layers of the KAN, $n_j$ are the number of nodes of the $j$-th layer, and $\phi_{i,j,k}$ are the activation functions. The original choice of $\phi$ proposed by \cite{liu2024kan} were B-splines of order $k$, which is what is used in this work, but other basis have been explored, such as wavelets \cite{bozorgasl2024wav}, radial basis functions \cite{li2024kolmogorov} and  Chebyshev \cite{ss2024chebyshev} polynomials. This structure imparts KANs with parameter efficiency and interpretability, especially for functions exhibiting compositional structures, outperforming traditional MLPs in these scenarios \cite{shukla2024comprehensive}.

\subsection{Physics-Informed Kolmogorov-Arnold Networks (PIKANs)}

In PIKANs, we seek to approximate the solution of a general initial-boundary value problem (IBVP), consisting of a partial differential equation (PDE) defined over a spatial domain $\Omega$ and a temporal domain $\Gamma$, along with corresponding initial and boundary conditions. Formally, the IBVP can be expressed as:
\begin{align}
    \mathcal{D}\left[u(\boldsymbol{x}, t)\right] &= f(\boldsymbol{x}, t), \quad \boldsymbol{x} \in \Omega, \; t \in \Gamma, \label{eq:general_pde} \\
    u(\boldsymbol{x}, 0) &= u_0(\boldsymbol{x}), \quad \boldsymbol{x} \in \Omega, \label{eq:initial_condition} \\
    u(\boldsymbol{x}, t) &= g(\boldsymbol{x}, t), \quad \boldsymbol{x} \in \partial\Omega, \; t \in \Gamma, \label{eq:boundary_condition}
\end{align}

where $\mathcal{D}$ represents the differential operator defining the PDE, $u(\boldsymbol{x}, t)$ is the solution, $f(\boldsymbol{x}, t)$ is the source term, $u_0(\boldsymbol{x})$ denotes the initial condition at time $t=0$, and $g(\boldsymbol{x}, t)$ specifies the boundary conditions on the boundary $\partial\Omega$ of the spatial domain.

Using the KAN approximation as defined in Equation~\eqref{eq:kan_approximation}, we seek to find an approximate solution $\hat{u}$ by minimizing the physics-informed loss functions. These loss functions, introduced by \cite{raissi2019physics}, are constructed as three distinct residuals obtained by summing over the interior, boundary, and initial collocation points of the domain. The residuals, denoted as $\mathcal{L}_{\text{pde}}$, $\mathcal{L}_{\text{ic}}$, and $\mathcal{L}_{\text{bc}}$, are defined as:
\begin{align}
    \mathcal{L}_{\text{pde}}(\theta) &= \frac{1}{N_{\text{pde}}} \sum_{i=1}^{N_{\text{pde}}} \left|\mathcal{D}\left[\hat u\left(\boldsymbol{x}_{\text{pde}}^i, t_{\text{pde}}^i ; \; \theta \right)\right] - f\left(\boldsymbol{x}_{\text{pde}}^i, t_{\text{pde}}^i\right) \right|^2,\label{eq:lpde} \\
    \mathcal{L}_{\text{ic}}(\theta) &= \frac{1}{N_{\text{ic}}} \sum_{i=1}^{N_{\text{ic}}} \left| \hat u\left(\boldsymbol{x}_{\text{ic}}^i, 0; \; \theta\right) - u_0\left(\boldsymbol{x}_{\text{ic}}^i\right) \right|^2,\label{eq:lic} \\
    \mathcal{L}_{\text{bc}}(\theta) &= \frac{1}{N_{\text{bc}}} \sum_{i=1}^{N_{\text{bc}}} \left| \hat u\left(\boldsymbol{x}_{\text{bc}}^i, t_{\text{bc}}^i; \; \theta\right) - u_{\text{bc}}\left(\boldsymbol{x}_{\text{bc}}^i, t_{\text{bc}}^i\right) \right|^2 \label{eq:lbc}
\end{align}
where $\theta$ is the set of trainable parameters of the KAN. The total loss function $\mathcal{L}$ used for training the PIKAN is constructed as a weighted sum of these individual loss components:
\begin{equation}
    \mathcal{L}(\theta) = \lambda_{\text{pde}}\mathcal{L}_{\text{pde}}(\theta) + \lambda_{\text{ic}} \mathcal{L}_{\text{ic}}(\theta) + \lambda_{\text{bc}} \mathcal{L}_{\text{bc}}(\theta),
    \label{eq:total_loss}
\end{equation}
where the weighting parameters can be adjusted manually or in a self-adaptive manner (see, \emph{e.g.}, \cite{mcclenny2023self,anagnostopoulos2024residual,chen2024self}).

For forward problems, we seek to find the optimal network parameters $\theta^\ast$ that minimize the total loss function $\mathcal{L}(\theta)$, ensuring that the predicted solution $\hat{u}(\boldsymbol{x}, t; \theta^\ast)$ satisfies the governing PDE, initial conditions, and boundary conditions. This optimization problem can be formally expressed as:

\begin{equation}
    \theta^\ast = \underset{\theta}{\arg\min}\; \mathcal{L}(\theta).
    \label{eq:minimize_loss}
\end{equation}

By solving the minimization problem in Equation~\eqref{eq:minimize_loss}, we obtain the network parameters that yield the approximate solution $\hat{u} \approx u$. The optimization process is typically performed using gradient-based methods, where the gradients of the loss function with respect to the network parameters are computed via automatic differentiation.

\subsection{Separable Physics-Informed Kolmogorov-Arnold Networks}

In the original construction of the KAN approximation, as presented in Equation~\eqref{eq:kan_approximation}, all input variables are processed jointly within a single network architecture. Specifically, a KAN approximates a multivariate function $u: \mathbb{R}^n \to \mathbb{R}$, where $u(\boldsymbol{x}) = u(x_1, x_2, \dots, x_n)$, by considering all variables simultaneously within a composite functional mapping. While this approach is effective for low-dimensional problems, it becomes computationally prohibitive for high-dimensional cases due to the exponential growth in the number of required collocation points and the increased complexity of the network architecture—a manifestation of the so-called curse of dimensionality.

One way to mitigate this is by decomposing the multivariate function $u(\boldsymbol{x}, t)$ into a separable representation. This idea, introduced by \cite{cho2024separable} for MLP-based PINNs, allows for a significant reduction in memory and computational cost, and can be seen as a tensor factorization. In the present work, we introduce the separable formulation for KANs: specifically, we approximate the solution $u(\boldsymbol{x}, t)$ as a finite sum of products of univariate functions, each depending on a single input variable:

\begin{equation}
    u(\boldsymbol{x}, t) \approx \hat{u}(\boldsymbol{x}, t) = \sum_{j=1}^{r} \prod_{i=1}^{n} f_j^{(\theta_i)}(x_i),
    \label{eq:separable_approximation}
\end{equation}

where $r \in \mathbb{N}$ is the rank of the approximation (also referred to as the latent dimension), $n$ is the number of spatial dimensions, $f_j^{(\theta_i)}: \mathbb{R} \to \mathbb{R}$ are univariate functions parameterized by $\theta_i$, and $x_i$ represents the $i$-th input variable. The time variable $t$ can be included as an additional dimension.

By adopting this separable representation, we effectively decompose the multivariate function into a sum over $r$ terms, each of which is a product of univariate functions. Each univariate function $f^{(\theta_i)}(x_i)$ can be approximated using a separate KAN or another suitable univariate approximator. This decomposition transforms the high-dimensional approximation problem into multiple one-dimensional problems, significantly reducing computational complexity.

To incorporate this separable approximation into the PIKAN framework, we adjust the physics-informed loss functions accordingly. The predicted solution $\hat{u}(\boldsymbol{x}, t)$ is defined as per Eq.~\eqref{eq:separable_approximation}. The partial derivatives of $\hat{u}$ with respect to each input variable $x_i$ are computed using the product rule:

\begin{equation} \frac{\partial \hat{u}}{\partial x_i} = \sum_{j=1}^{r} \left( f_j^{(\theta_i)'}(x_i) \prod_{\substack{k=1 \ k \neq i}}^{n} f_j^{(\theta_k)}(x_k) \right), \label{eq:separable_derivative} \end{equation}

where $f_j^{(\theta_i)'}(x_i)$ denotes the derivative of $f_j^{(\theta_i)}$ with respect to $x_i$. Higher-order derivatives can be computed similarly by applying the product rule iteratively. We also define $f^{(\theta_i)}=(f_1^{(\theta_i)},\ldots,f_r^{(\theta_i)}).$

A key advantage of the separable representation in SPIKANs is that each KAN approximates a univariate function $f^{(\theta_i)}: \mathbb{R} \to \mathbb{R}^r$. This design enables the efficient computation of the derivatives required to apply the differential operator $\mathcal{D}$ to the approximation $\hat{u}$, as shown in Eq.~\eqref{eq:separable_derivative}, by leveraging forward-mode automatic differentiation. Forward-mode is particularly effective for functions with a single input and multiple outputs, which aligns perfectly with the structure of each $f^{(\theta_i)}$ in the separable framework. In contrast, reverse-mode automatic differentiation is optimized for functions that have many inputs and a single output and is commonly utilized in traditional PIKANs. For a more detailed discussion on this topic, see \cite{cho2024separable}.

\begin{figure}[h]
    \centering
    \includegraphics[width=1.0\textwidth]{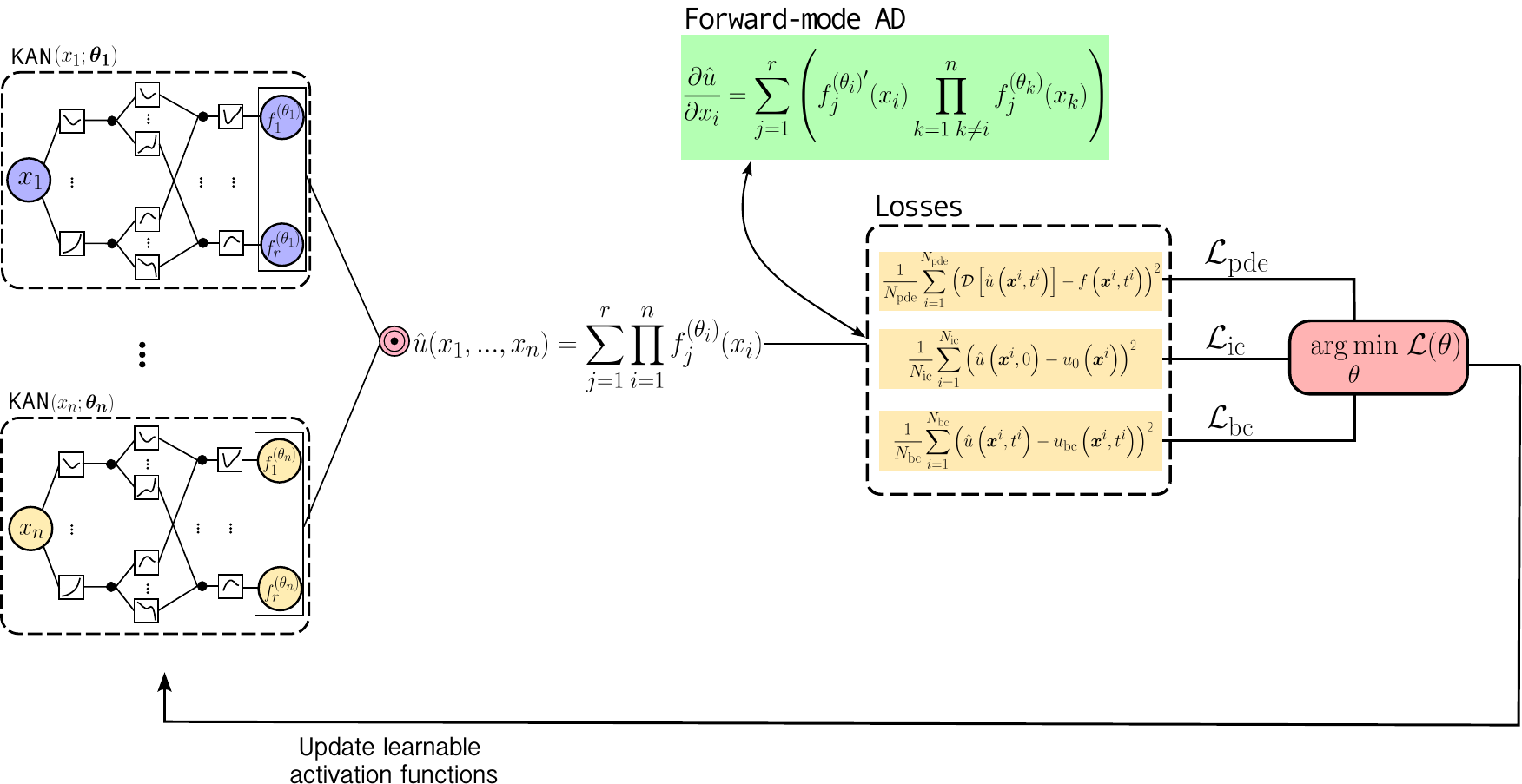}
    \caption{Schematic of the SPIKAN architecture. We decompose the $n$-dimensional PDE into $n$ individual KANs. In contrast with PIKANs, which take as input points given by tuples $(x_1, ..., x_n)$, in SPIKANs, each network receives only one coordinate $x_i$ at a time. Each network produces $r$ outputs, which are then combined via an outer product and summation on the latent dimension (denoted by the $\odot$ operator) to approximate the field $\hat{u}(x_1,..., x_n)$. We use forward-mode automatic differentiation to evaluate the derivatives of $\hat{u}$ needed to satisfy the differential operator $\mathcal{D}$ and any other derivative needed to construct the losses. The weights of the networks are then adjusted via gradient descent.}
    \label{fig:separable_pikan_diagram}
\end{figure}

The approximation of the solution $u$ follows the same construction of the PINN and PIKAN loss, \emph{i.e.}, Eqs.~\eqref{eq:lpde}-\eqref{eq:lbc}. A diagram illustrating the full architecture of the proposed SPIKAN method is shown in Fig.~\ref{fig:separable_pikan_diagram}.

\section{Results}
\label{sec:results}

In this section, we present a validation study of the proposed method, as well as comparisons with the original implementation of PIKANs \cite{shukla2024comprehensive}. All the networks in this work, including the base PIKANs, are trained with Adam optimizer, with an initial learning rate $l=10^{-3}$ and hyperparameters $\beta_1=0.9$ and $\beta_2 = 0.999$. The total number of epochs used for training is specified in each subsection. All PIKANs and SPIKANs tested used a constant KAN grid size of $g = 3$. We leave the analysis of the multigrid aspect of KANs and PIKANs \cite{rigas2024adaptive} via variable $g$ for future work. We use $k$ to denote the degrees of the B-splines in both PIKANs and SPIKANs.

For PIKANs, the size of the network is described as $[w_{in}, w_{h_1}, ...,w_{h_n}, w_{out}]$, where $w_{in}, w_{h_j}, w_{out}$ denotes the width of the input layer, $j$-th hidden layer and output layers, respectively. Similarly, for SPIKANs, a network of size $d \times [w_{in} = 1, w_{h_1}, ...,w_{h_n}, r*w_{out}]$ denotes width of input, hidden and output layers, with additional hyperparameters $d$ and $r$ indicating the dimensionality of the PDE (total number of networks used) and the latent dimension, respectively. Notice that for SPIKANs, $w_{in} = 1$ by construction (c.f. Fig.~\ref{fig:separable_pikan_diagram}).

Throughout this section, for both PIKANs and SPIKANs, we have used an unbiased loss function, where $\lambda_\text{ic}= \lambda_\text{bc} = \lambda_\text{pde} = 1$. $L_2$ errors are measured as:

\begin{equation}
    L_2 = \sqrt{\frac{\sum_{i=1}^{N_\text{cp}} \left| u^{\text{ref}}(\boldsymbol{x}_i) - \hat{u}(\boldsymbol{x}_i) \right|^2}{\sum_{i=1}^{N_\text{cp}} \left| u^{\text{ref}}(\boldsymbol{x}_i) \right|^2}},
\end{equation}

where $u^\text{ref}$ denotes the reference solution, $\hat{u}$ is the approximate solution and $N_\text{cp}$ is the number of collocation points.

The implementations used in this work are based on \texttt{Jax}~\cite{jax2018github}, using the \texttt{Jax-KAN} package \cite{Rigas_jaxKAN_A_JAX-based_2024, rigas2024adaptive} for the KAN implementation. All simulations were performed using an Nvidia T4 Tensor Core GPU with 16GB of vRAM.

\subsection{2D Helmholtz equation}

Consider the following boundary value problem for the 2D Helmholtz equation:

\begin{equation}
    \frac{\partial^2 u}{\partial x^2} + \frac{\partial^2 u}{\partial y^2} + \kappa^2u = q(x, y), \quad (x,y) \in \Omega, 
\end{equation}

subject to the homogeneous Dirichlet boundary conditions:
\begin{align}
    u(-1, y) &= 0, \quad -1 \leq y \leq 1 \\
    u(1, y) &= 0, \quad -1 \leq y \leq 1 \\
    u(x, -1) &= 0, \quad -1 \leq x \leq 1 \\
    u(x, 1) &= 0, \quad -1 \leq x \leq 1
\end{align}

where $\Omega = \{(x, y) \in [-1, 1]^2\}$. The forcing term $q(x,y)$ is given by:

\begin{equation}
    \begin{split}
        q(x,y) = &-(a_1\pi)^2 \sin(a_1\pi x) \sin(a_2\pi y) + \sin(a_1\pi x) \sin(a_2\pi y) \\
                 &- (a_2\pi)^2 \sin(a_1\pi x) \sin(a_2\pi y) + \kappa^2 \sin(a_1\pi x) \sin(a_2\pi y),
    \end{split}
\end{equation}

which is obtained via the manufactured solution

\begin{equation}
    u(x,y) = \sin(a_1 \pi x) \sin(a_2 \pi y).
    \label{eq:analytical_2dhelmholtz}
\end{equation}
 
We consider the case $\kappa = 1$, $a_1 = 1$ and $a_2=4$, as in \cite{shukla2024comprehensive}. 
 
Due to the difference in the architecture of PIKANs and SPIKANs, we perform tests with different number of neurons in the hidden layers, but focusing on a similar number of trainable parameters. Since the problem is in two dimensions, in the SPIKAN cases, the reported KAN sizes refer to the two networks that compose the SPIKAN, and the last size corresponds to the latent dimension, set to $r=5$ for all the cases. For both PIKANs and SPIKANs, we use B-splines of degree $k=3$. The optimization was performed using the Adam optimizer with initial learning rate $10^{-3}$ for 20,000 epochs. The collocation points are arranged in an equidistant manner, forming a grid of $n_\text{cp} = n_x \times n_y$ collocation points in the $\Omega$ domain.

Table \ref{tab:2dhelmholtz} summarizes the results of all tested cases. The iteration times were computed as an average over all the epochs, and were all produced in the same hardware environment. We note that SPIKANs achieve speedups ranging from $O(10)$ to $O(100)$, by virtue of the reduced number of network evaluations from the now one-dimensional inputs. The calculations of the $L_2$ error are performed with respect to the analytical solution, Eq.~\eqref{eq:analytical_2dhelmholtz}. 

While the advantages of SPIKANs lie in decreasing the computational intesity of training, SPIKANs notably also achieved superior accuracy for all the tested cases, with significant decreases in computational time. For the PIKAN case (c), we noticed a deterioration in the quality of the solution, possibly due to the fixed number of epochs used for all the networks. This could be caused by the fixed number of epochs used in all the examples; larger PIKANs, exposed to larger training datasets, could require longer to converge.

It is important to highlight that the reduction in computational cost does not come at the cost of an increase in the memory; on the contrary, memory utilization is reduced from $O(n_x * n_y + \text{No. Parameters})$ to roughly $O(d*\sqrt{n_x * n_y} + \text{No. Parameters})$, with $d=2$ for the 2D Helmholtz example. For a detailed description of the memory utilization in separable networks, we refer the reader to \cite{cho2024separable}. 

\begin{figure}[h!]
    \centering
    \includegraphics[width=1.0\linewidth]{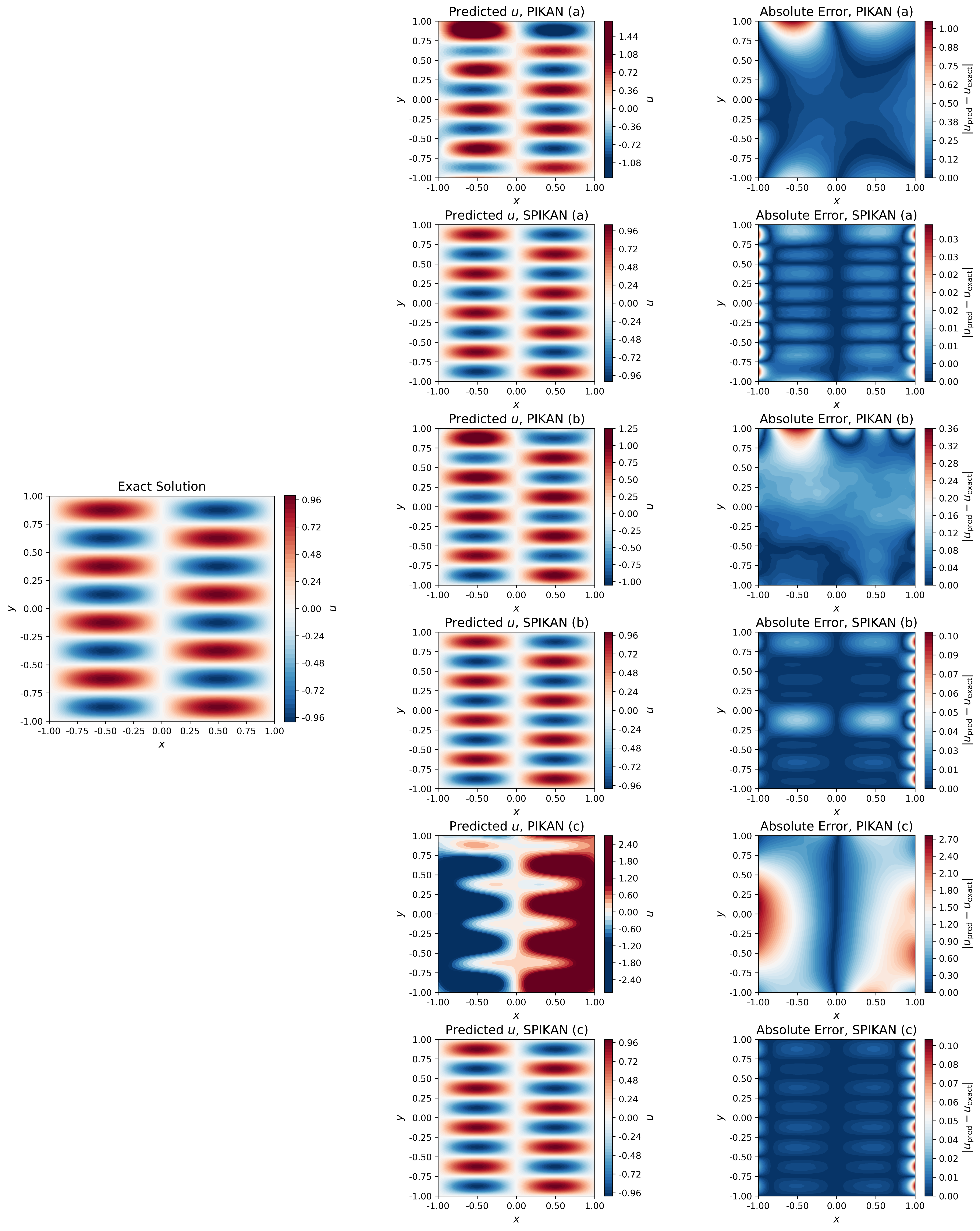}
    \caption{Contour plots and absolute error between the predicted solution $u$ obtained with PIKANs, SPIKANs and the analytical solution of the 2D Helmholtz boundary value problem. A description of the cases is shown in Table~\ref{tab:2dhelmholtz}.}
    \label{fig:2d_helmholtz_separable_pikan}
\end{figure}

\begin{table}[h!]
    \centering
    \begin{tabular}{|l|c|c|c|c|c|c|}
        \hline
        \textbf{Method} & \textbf{KAN Size}  & $n_\text{cp}$  & \textbf{No. Parameters}& \textbf{$L_2$}(\%) & \textbf{Time (ms/iter)} & \textbf{Speedup} \\
        \hline\hline
        PIKAN (a) & [2,6,6,1] & $100^2$  & 445 & 37.26 & 20.75 & 1 (baseline)\\ 
        \hline
        SPIKAN (a) & $2 \times [1,3,3,1*5]$ & $100^2$ & 454 & 1.29 & 2.65  & 7.8\\
        \hline\hline
        PIKAN (b) & [2,9,9,1] & $100^2$ & 883 & 15.73 & 132.63 & 1 (baseline) \\ 
        \hline
        SPIKAN (b) & $2 \times [1,5,5,1*5]$ & $100^2$ & 910 & 2.91 & 2.73  & 48.6  \\
        \hline\hline
        PIKAN (c) & [2,9,9,1] & $200^2$ & 883 & 228.84 & 961.54 & 1 (baseline) \\ 
        \hline
        SPIKAN (c) & $2 \times [1,5,5,1*5]$ & $200^2$ & 910 & 2.20 & 3.35 &  287.0 \\
        \hline\hline
    \end{tabular}
    \caption{Comparison of methods for the 2D Helmholtz boundary value problem.}
    \label{tab:2dhelmholtz}
\end{table}

\subsection{2D steady lid-driven cavity flow}

In this section, we demonstrate usage of SPIKANs in a multiple output setup: we solve the 2D steady, incompressible Navier-Stokes equations in an enclosed, lid-driven cavity. In this benchmark, the flow is driven by a constant tangential velocity applied to the lid of the cavity. The governing equations are given by

\begin{align}
    \frac{\partial u}{\partial x} + \frac{\partial v}{\partial y} &= 0, \\
    u \frac{\partial u}{\partial x} + v \frac{\partial u}{\partial y} &= - \frac{\partial p}{\partial x} + \frac{1}{Re} \left( \frac{\partial^2 u}{\partial x^2} + \frac{\partial^2 u}{\partial y^2} \right), \\
    u \frac{\partial v}{\partial x} + v \frac{\partial v}{\partial y} &= - \frac{\partial p}{\partial y} + \frac{1}{Re} \left( \frac{\partial^2 v}{\partial x^2} + \frac{\partial^2 v}{\partial y^2} \right),
\end{align}

where $(x, y) \in \Omega = [0, 1]^2$, $u$ and $v$ are the velocity components in the $x$ and $y$-directions, respectively, and $p$ is the pressure field. Here, $Re$ denotes the Reynolds number, which characterizes the ratio of inertial forces to viscous forces in the flow. In the example below, we consider $Re = 100$.

The boundary conditions are specified as follows:

\begin{align}
    u(x, 1) &= 1, \quad v(x, 1) = 0, \quad 0 \leq x \leq 1 \quad \text{(top boundary, moving lid)}, \\
    u(x, 0) &= 0, \quad v(x, 0) = 0, \quad 0 \leq x \leq 1 \quad \text{(bottom boundary, no-slip)}, \\
    u(0, y) &= 0, \quad v(0, y) = 0, \quad 0 \leq y \leq 1 \quad \text{(left boundary, no-slip)}, \\
    u(1, y) &= 0, \quad v(1, y) = 0, \quad 0 \leq y \leq 1 \quad \text{(right boundary, no-slip)}.
\end{align}

Similarly to the 2D Helmholtz example, the optimization was performed using the Adam optimizer with initial learning rate $10^{-3}$. Training is performed in 50,000 epochs, with B-splines of degree $k=5$ for both PIKAN and SPIKANs. The collocation points are arranged in an equidistant manner, forming a grid of $n_\text{cp} = n_x \times n_y$ points, uniformly distributed within the $\Omega$ domain.

The reference fields were obtained numerically, using the finite volume method (FVM). The steady-state SIMPLE \cite{patankar1983calculation} algorithm was employed to resolve the incompressible Navier-Stokes equations in a equally spaced 256x256 cell mesh. Convective fluxes were approximated with a first-order upwind scheme to maintain stability, while the diffusive terms were discretized using a second-order Gauss linear scheme. A pressure reference was specified to ensure convergence, with a target tolerance of $10^{-4}$ for pressure and $10^{-5}$ for the velocity components. The iterative solver configuration utilized the Preconditioned Conjugate Gradient (PCG) method for the pressure field and a Gauss-Seidel smoother for the velocity field, with relaxation factors of $0.3$ for pressure and $0.7$ for velocity to enhance convergence stability. In addition, we compare the vertical and horizontal centerline velocity profiles with the benchmark results of \cite{ghia1982high}.

\begin{table}[h!]
    \centering
    \begin{tabular}{|l|c|c|c|c|c|c|}
        \hline
         \textbf{Method} & \textbf{KAN Size}  & $n_\text{cp}$  & \textbf{No. Parameters} & $L_2(u,v,p)$ (\%)& \textbf{Time (ms/iter)} & \textbf{Speedup} \\
        \hline\hline
        PIKAN & [2,9,9,3] & $50^2$  & 1029 & (9.73, 11.80, 48.71) & 301.20 & 1 (baseline)\\ 
        \hline
        SPIKAN (a) & $2 \times [1,5,5,3*5]$ & $50^2$ & 2150  & (8.66, 12.01, 35.91) & 4.3 & 70.05 
        \\
        \hline
        SPIKAN (b) & $2 \times [1,5,5,3*5]$ & $100^2$ & 2150 & (5.83, 7.18, 29.55) & 6.14  & 49.05 \\
        \hline\hline
    \end{tabular}
\caption{Comparison of methods for the 2D steady Navier-Stokes (lid-driven cavity) for $Re=100$.}
\label{tab:cavity_results}
\end{table}

Table~\ref{tab:cavity_results} summarizes the results. We use a PIKAN with architecture of shape [2,9,9,3] as a baseline. The $L_2$ differences are compared with the FVM results as reference. We normalize the pressure field $p$ by removing its mean and scaling it with the maximum value, in order to achieve a translationally invariant field across all the methods tested.

\begin{figure}[!h]
    \centering
    \includegraphics[width=1.0\linewidth]{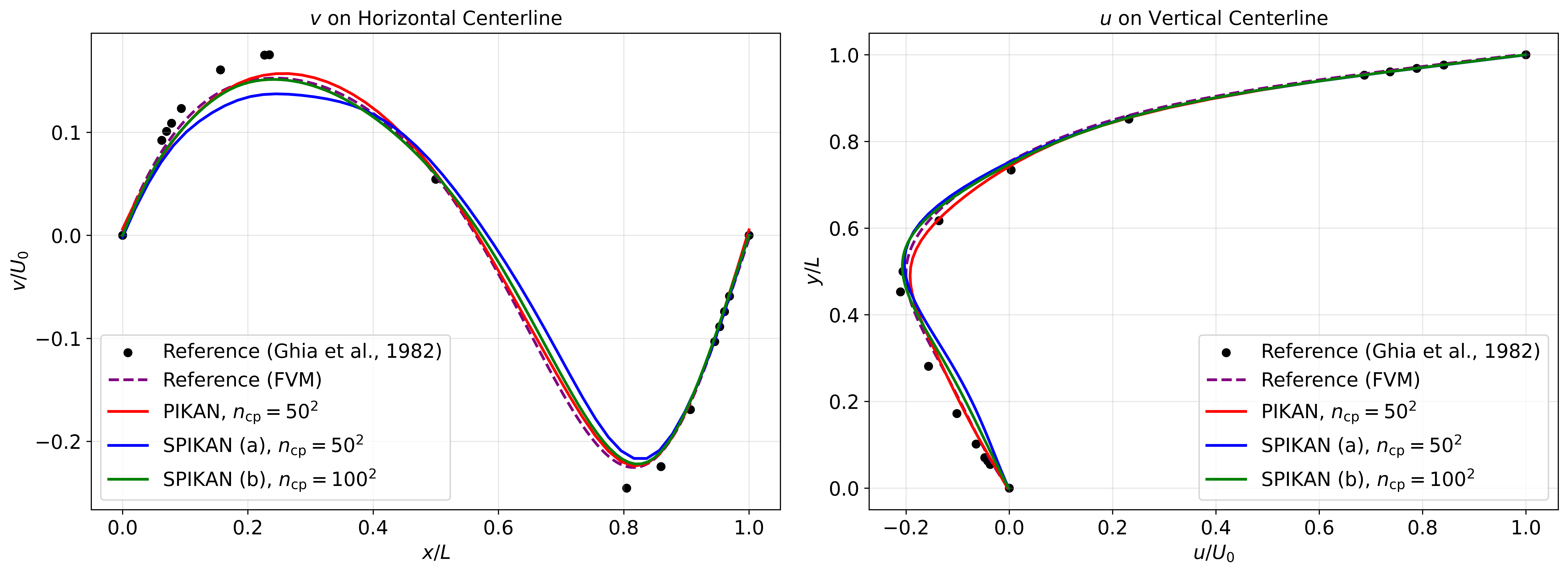}
    \caption{Comparison of horizontal and vertical velocity profiles obtained with PIKAN ($[2,5,5,3], n_\text{cp}=50^2)$) and SPIKANs ($2 \times [1,9,9,3*5], n_\text{cp}=50^2, 100^2$) with the reference \cite{ghia1982high} for $Re = 100$.}
    \label{fig:cavity_velocity_profiles_Re100}
\end{figure}

\begin{figure}[!h]
    \centering
    \includegraphics[width=1.0\linewidth]{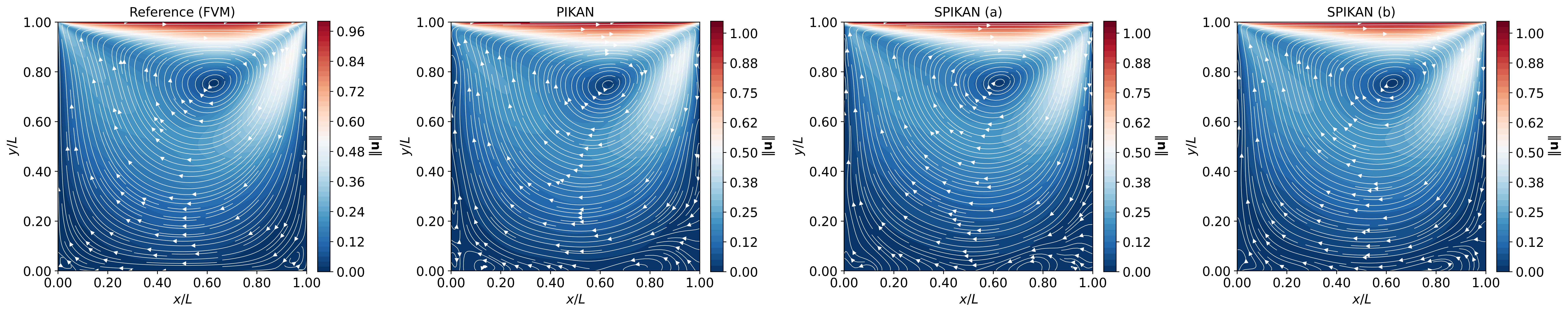}
    \caption{Comparison of the isocontours of $||\mathbf{u}|| = \sqrt{u^2 + v^2}$ and streamlines, obtained with finite volume method, PIKAN ($[2,9,9,3], n_\text{cp}=50^2)$), SPIKAN (a) ($2 \times [1,5,5,3*5], n_\text{cp}=50^2$) and SPIKAN (b) ($2 \times [1,5,5,3*5], n_\text{cp}=100^2$). }
    \label{fig:cavity_comparison_fields_Re100}
\end{figure}

\begin{figure}[!h]
    \centering
    \includegraphics[width=1.0\linewidth]{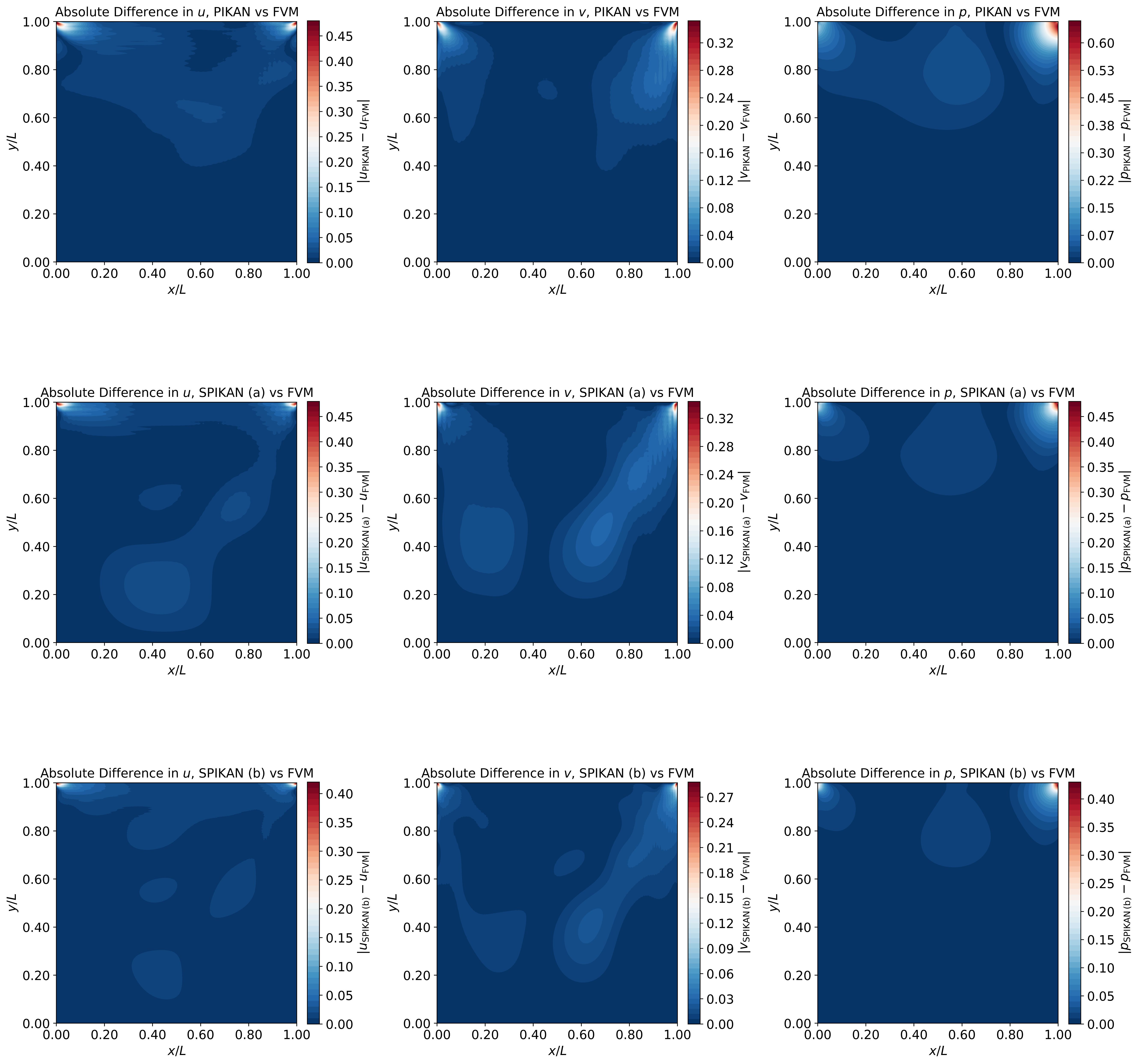}
    \caption{Comparison of absolute difference of predicted $u, v$ and $p$ fields obtained from PIKAN ($[2,9,9,3], n_\text{cp}=50^2$) and SPIKAN ($2 \times [1,5,5,3*5], n_\text{cp}=50^2$) versus the results obtained with FVM for $Re = 100$.}
    \label{fig:absolute_difference_comparison}
\end{figure}

Fig.~\ref{fig:cavity_velocity_profiles_Re100} shows the a comparison of the centerline velocity profiles of the baseline PIKAN ($[2,9,9,3], n_\text{cp}=50^2$) and SPIKANs (a) ($2 \times [1,5,5,3*5], n_\text{cp}=50^2$) and (b) ($2 \times [1,5,5,3*5], n_\text{cp}=100^2$). Despite all results being close in terms of accuracy, the SPIKAN (b) outperforms the other cases, with the lowest $L_2$. Increasing the number of collocation points for training from $n_\text{cp} = 50^2$ to $n_\text{cp} = 100^2$ brings the velocity profiles closer to the references. Even though this might seem an obvious result, we highlight that doubling the number of collocation points in each dimension would lead to a quadruple ($O(2^d)$) increase in computational time in PIKANs, but only a linear increase in SPIKANs. This effect becomes more prominent in higher dimensions, as we demonstrate in Sec.~\ref{sec:2d1+1_klein_gordon}. 

Another important result is that SPIKAN (a) achieved similar accuracy of the baseline PIKAN with a $~70\times$ speedup. Even with a finer resolution of $n_\text{cp} = 100^2$, SPIKAN (b) still outperformed PIKAN and trained nearly $50\times$ faster.

Contour plots of the velocity magnitude $||\mathbf{u}||$, along with the streamlines of the FVM reference, PIKAN and SPIKANs (a) and (b) are shown in Fig.~\ref{fig:cavity_comparison_fields_Re100}. Despite having streamlines crossing the wall boundaries, notice that the SPIKAN predictions were far closer to the reference than PIKANs, especially in the lateral walls, $x=0$ and $x=1$. This nonphysical behavior is a consequence of using a weak formulation of boundary conditions, which in turn lead to larger errors near the boundaries. Fig.~\ref{fig:absolute_difference_comparison} further highlights this, by showing the absolute difference of fields $u$, $v$ and $p$ with the reference FVM counterpart.

\subsection{1D+1 Allen-Cahn equation}
In this example, we test the proposed architecture in what has been shown to be a challenging case for PINNs and PIKANs: the Allen-Cahn equation. This nonlinear PDE describes the reaction-diffusion in phase separation, and is widely used in studies of alloy separation. In its one-dimensional form, the equation is given by

\begin{equation}
    \frac{\partial u}{\partial t} - D \frac{\partial^2 u}{\partial x^2} + 5(u^3-u) = 0.
    \label{eq:ac_eqn}
\end{equation}

We examine the case with $D= 10^{-4}$, $(x,t) \in [0,1]\times[-1,1]$. The initial and boundary conditions are the same ones used by \cite{shukla2024comprehensive}:

\begin{align}
    u(x, 0)  &= x^2 \cos(\pi x), \label{eq:ac_ic} \\
    u(-1, t) &= u(1,t) = -1.
    \label{eq:ac_bcs}
\end{align}

Previous studies have reported training difficulties for this system in the context of PINNs \cite{wight2020solving, mattey2022novel} and PIKANs \cite{shukla2024comprehensive}. These difficulties have been attributed to the presence of a non-trivial fixed point, which typically correspond to a global minima of the physics loss term $\mathcal{L}_\text{PDE}$ shown in Eq.~\eqref{eq:lpde} with non-trivial basins of attraction \cite{rohrhofer2022role}. As a consequence, the solution can only be learned with specific choices of $\lambda_\text{ic}, \lambda_\text{bc}$, such that $\lambda_\text{ic}, \lambda_\text{bc} \gg \lambda_\text{pde}$ (c.f. Eq.\eqref{eq:total_loss}), or under variations in architecture (e.g., via multi-fidelity learning, as in \cite{howard2024multifidelity, heinlein2024multifidelity}) or adaptive weighting strategies \cite{anagnostopoulos2024residual}. For the sake of clarity, we limit the scope of the hyperparameter space and focus on the case of an unbiased loss function, where $\lambda_\text{ic}= \lambda_\text{bc} = \lambda_\text{pde} = 1$. This choice allows for a fair comparison of the proposed method.

In this section, we compare the solutions obtained via PIKAN and SPIKAN training, focusing on the effect of the latent dimension $r$ in the SPIKAN architecture in the accuracy of the solution. A summary of the cases tested is shown in Table \ref{tab:allen_cahn_results}. The reference result, used to compute the $L_2$ and absolute errors, is obtained by solving the initial-boundary value problem Eqs.~\eqref{eq:ac_eqn}-\eqref{eq:ac_bcs} numerically using a Fourier pseudospectral method in space and a 4th order Runge-Kutta method in time, on a $n_x = 320, n_t = 1000$ resolution.

Table \ref{tab:allen_cahn_results} summarizes the results. As a baseline, we use a PIKAN of size [2,9,9,1], with B-splines of degree $k = 5$ and $n_\text{cp} = 320 \times 160$ collocation points. We explore three SPIKAN cases, all with $k = 5$ and $n_\text{cp} = 320 \times 160$, but with increasing values of $r$. For all the cases, we used the Adam optimizer with initial learning rate of $10^{-3}$ for 150,000 epochs. 

\begin{table}[h!]
    \centering
    \begin{tabular}{|l|c|c|c|c|c|c|}
        \hline
        \textbf{Method} &\textbf{KAN Size} & $n_\text{cp}$  & \textbf{No. Parameters}& \textbf{$L_2$} (\%) & \textbf{Time (ms/iter)} & \textbf{Speedup} \\
        \hline\hline 
        PIKAN & [2,9,9,1]  & $320\times160$  & 1099 & 52.92 & 454.54 & 1 (baseline)\\ 
        \hline
        SPIKAN (a)& $2 \times [1,5,5,1*5]$  & $320\times160$ & 1130 & 94.18  & 3.53 & 128.76 \\
        \hline
        SPIKAN (b)& $2 \times [1,5,5,1*10]$  & $640\times320$ & 1640 & 17.15 &4.36  & 104.25 \\
        \hline
        SPIKAN (c)& $2 \times [1,5,5,1*20]$  &  $320\times160$ & 2660 & 28.54 & 5.22 & 87.08 \\
        \hline\hline
    \end{tabular}
    \caption{Comparison of methods for the 1D+1 Allen-Cahn initial-boundary value problem.}
    \label{tab:allen_cahn_results}
\end{table}

\begin{figure}[!h]
    \centering
    \includegraphics[width=1.0\linewidth]{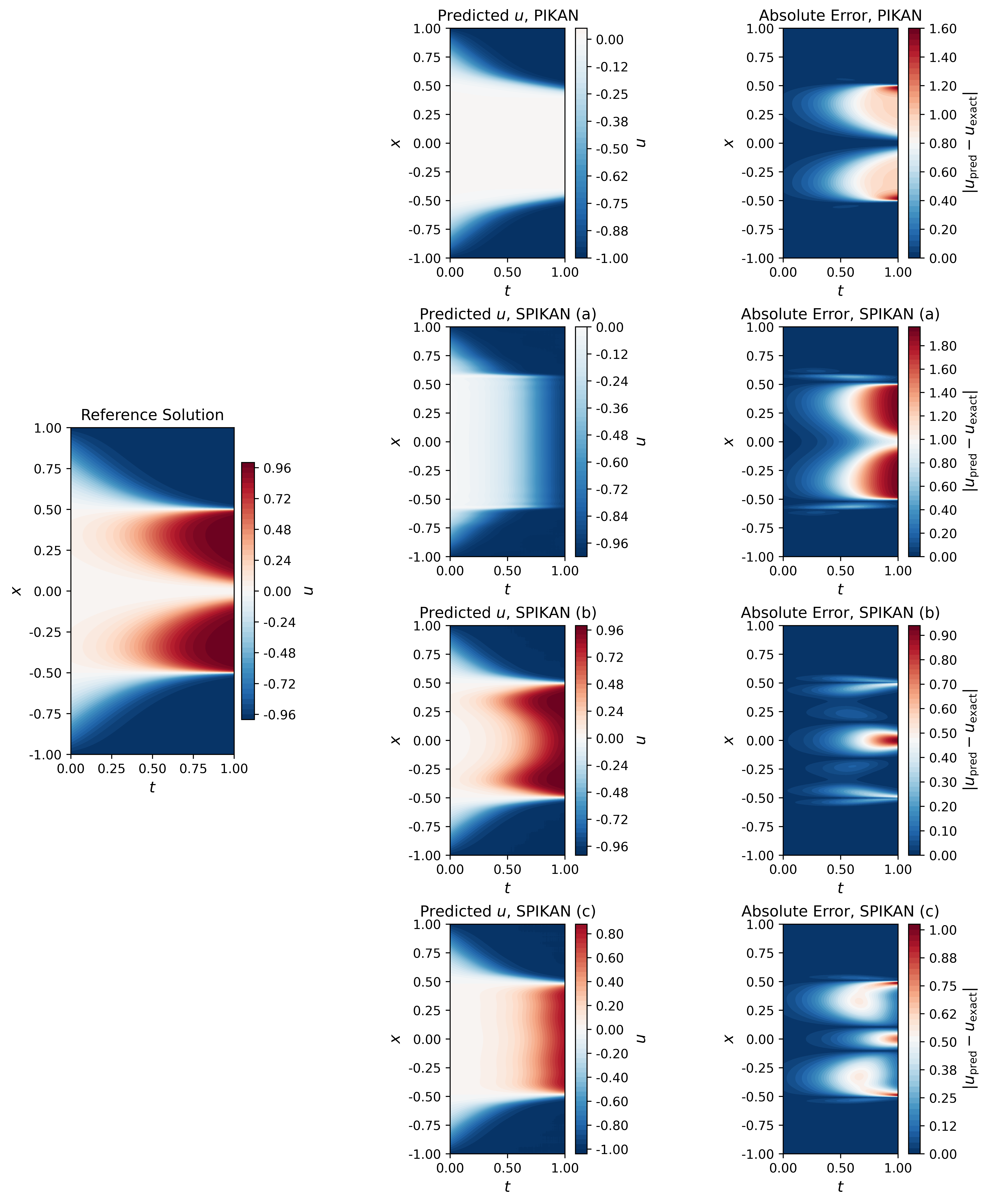}
    \caption{Isocontours of $u$ of the reference solution, PIKAN ([2,9,9,1]) and SPIKAN cases (a) $2 \times [1,5,5,1*5]$, (b) $2 \times [1,5,5,1*10]$ and (c) $2 \times [1,5,5,1*20]$,   along with absolute errors for the 1D+1 Allen-Cahn equation. Implementation details are shown in Table \ref{tab:allen_cahn_results}.}
    \label{fig:comparison_allen_cahn}
\end{figure}

Contour plots of $u$ for the reference solution and the aforementioned networks are shown in Fig.~\ref{fig:comparison_allen_cahn}. For the PIKAN solution, our results are in agreement with what was reported in \cite{shukla2024comprehensive}, in which $u$ did not converge to the reference solution. This behavior of $u<0$ is consistent with the local minima observed when the optimization converges to the fixed point, as described in \cite{rohrhofer2022role}. 
For the SPIKAN (a) tested, with $r=5$, we note a similar behavior as of the PIKAN predictions, although obtained with a nearly $70 \times$ speedup. For SPIKAN (b), the use of $r=10$ drastically improves the accuracy of the results, allowing the correct range of $u$ and reducing the absolute errors in the center of the spatial domain ($x \in [-0.5,0.5]$ region). Despite the improvement for SPIKAN (b), the increase in latent dimension to $r = 20$ in SPIKAN (c) decreases the accuracy of the predictions, which can be caused by a slower learning process given that $r$ increases the overall number of trainable parameters in the network; given that all the cases used a fixed number of iterations, it is possible that this effect would be reduced with more training epochs. We leave a comprehensive study of the impact of $r$ in the stages of learning \cite{anagnostopoulos2024residual} for future work.

\subsection{2D+1 Klein-Gordon equation}
\label{sec:2d1+1_klein_gordon}
We extend our analysis to the time-dependent, inhomogeneous Klein-Gordon equation in a two-dimensional spatial domain. As in \cite{cho2024separable}, we use the governing equation:

\begin{equation}
    \frac{\partial^2 u}{\partial t^2} - \left( \frac{\partial^2 u}{\partial x^2} + \frac{\partial^2 u}{\partial y^2} \right) + u^2 = h(x, y, t), \quad (x,y) \in \Omega, \; t \in \Gamma,
    \label{eq:kg_pde}
\end{equation}

subject to the initial and boundary conditions:

\begin{align}
    u(x, y, 0) &= x_1 + x_2, \quad (x,y) \in \Omega, \label{eq:kg_ic} \\
    u(x, y, t) &= u_{\text{bc}}(x, y), \quad (x,y) \in \partial\Omega, \; t \in \Gamma, \label{eq:kg_bc}
\end{align}

where the spatial domain is defined as $\Omega = [0, 1]^2$ and the temporal domain as $\Gamma = [0, 10]$. For purpose of validation, a manufactured solution $u$

\begin{equation}
    u(x, y, t) = (x + y)\cos(t) + x y \sin(t),
    \label{eq:kg_exact}
\end{equation}

 is used to derive the forcing term $h(x, y, t)$ and the boundary condition $u_{\text{bc}}(x, y)$. For this specific choice of $u$, $h$ takes the form:

\begin{equation}
    h(x, y, t) = u^2 - u.
    \label{eq:kg_forcing}
\end{equation}

Table \ref{tab:klein_gordon_results} summarizes the results. The baseline network used for speedup calculations is a PIKAN of size [2,9,9,1], B-splines of degree $k = 3$, with $n_\text{cp} = 50^3$ collocation points. We note that above $~60^3$ collocation points, the memory requirement for PIKANs exceeds the vRAM capacity of the GPU used in this work. Such limitation can be alleviated with usage of larger GPUs or mini-batching strategies; the latter, however, comes with additional memory transfer overheads, which further increases the computational time needed for training.

To demonstrate the ability of collocation point refinement, we explore SPIKAN cases with larger values of $n_\text{cp}$. As in the PIKAN baseline, we set the degree of the B-splines $k=3$. For all the cases, we used the Adam optimizer with initial learning rate of $10^{-3}$ for 50,000 epochs.

\begin{table}[h!]
    \centering
    \begin{tabular}{|l|c|c|c|c|c|c|}
        \hline
        \textbf{Method} & \textbf{KAN Size}  & $n_\text{cp}$  & \textbf{No. Parameters}& \textbf{$L_2$} (\%) & \textbf{Time (ms/iter)} & \textbf{Speedup} \\
        \hline\hline
        PIKAN & [3,5,5,1] & $50^3$ & 371 & 2.16 & 847.45 & 1 (baseline)\\ 
        \hline
        SPIKAN (a) & $3 \times [1,3,3,1*10]$ & $100^3$ & 704 & 1.00 & 3.20 & 264.83 \\
        \hline
        SPIKAN (b) & $3 \times [1,5,5,1*10]$ & $150^3$ & 1320 & 0.76 & 5.98 & 141.71  \\
        \hline
        SPIKAN (c) & $3 \times [1,5,5,1*10]$ & $200^3$ & 1320 & 0.79 & 7.82 & 108.37 \\
        \hline\hline
    \end{tabular}
    \caption{Comparison of methods for the 2D+1 Klein-Gordon initial-boundary value problem.}
    \label{tab:klein_gordon_results}
\end{table}

\begin{figure}[!h]
    \centering
    \includegraphics[width=1.0\linewidth]{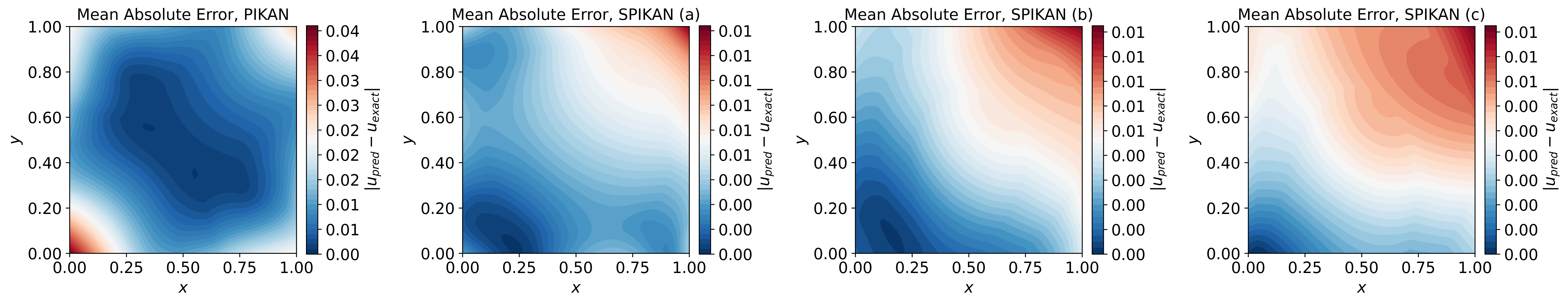}
    \caption{Isocontours of the mean absolute error, averaged over time, of the reference solution and PIKAN ([3,5,5,1]) and SPIKAN cases (a) $3 \times [1,3,3,1*10]$, (b) $3 \times [1,5,5,1*10]$ and (c) $3 \times [1,5,5,1*10]$, for different values of refinement $n_\text{cp}$, for the 2D+1 Klein-Gordon equation. Implementation details are shown in Table \ref{tab:klein_gordon_results}. Note the different scale of the color plots of the absolute error for PIKAN and SPIKAN.}
    \label{fig:comparison_klein_gordon}
\end{figure}

\begin{figure}[!h]
    \centering
    \includegraphics[width=0.6\linewidth]{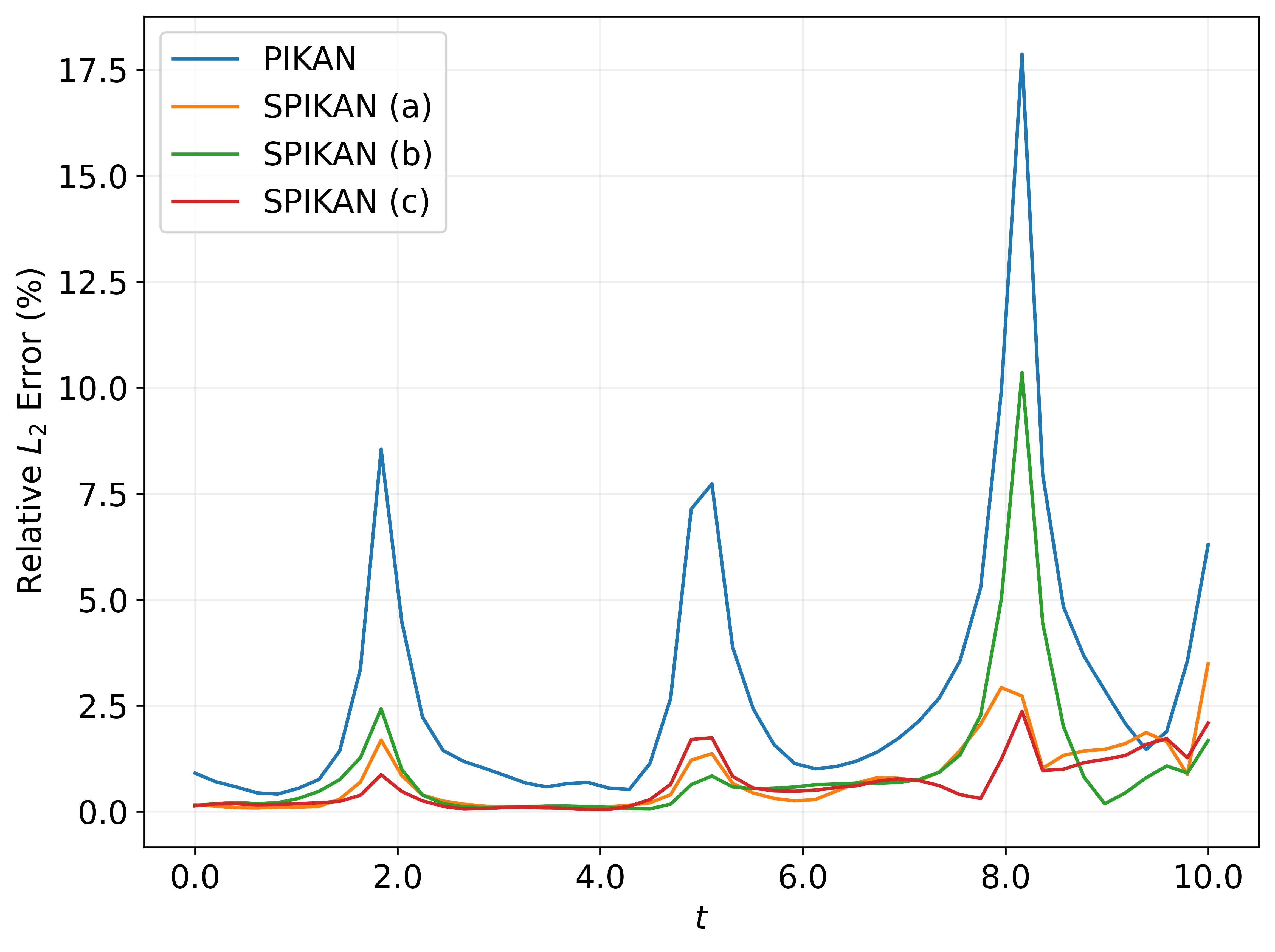}
    \caption{Temporal evolution of the $L_2$ error for the cases summarized in Table \ref{tab:klein_gordon_results}. SPIKAN predictions (a)-(c) achieved $O(100)$ speedup and yield results that are $50\%+$ more accurate in the $L_2$ norm versus the PIKAN baseline. The amplification of the periodic spikes reveal the difficulties associated with the lack of spatio-temporal causality, as discussed in \cite{wang2022respecting}.}
    \label{fig:l2_klein_gordon}
\end{figure}

Figure \ref{fig:comparison_klein_gordon} shows isocontours of the mean absolute errors, averaged over the time dimension. All the SPIKAN cases outperform the PIKAN baseline in accuracy and computational time. In this 3-dimensional setting, a speedup of $O(100)$ was obtained, even for SPIKAN (c), which has four times the number of trainable parameters of the baseline. The relative $L_2$ obtained in (c), however, shows a saturation of learning, indicating the need of larger KAN sizes to benefit of larger $n_\text{cp}$.

The temporal evolution of the relative $L_2$ is shown in Fig.~\ref{fig:l2_klein_gordon}. The instantaneous $L_2$ errors of the SPIKAN cases are lower than the PIKAN baseline. The periodicity of the curves reflects the choice of a periodic manufactured solution $u$; the spike formations that seem to grow over time, however, seem to be a consequence of the lack of spatio-temporal causality in the loss function formulation. A complete description of this limitation can be found in \cite{wang2022respecting}.

\section{Conclusions}
\label{sec:conclusions}

We have developed SPIKANs, an architecture for separable physics-informed Kolmogorov-Arnold networks. This method extends the work of \cite{cho2024separable} for KANs, significantly boosting the performance of these networks in both memory usage and computation time. Our tests demonstrate that SPIKAN predictions often yield more accurate results than those of PIKANs for a comparable number of trainable parameters, while providing speedups of $O(10)-O(100)$ in several benchmark tests.

\begin{figure}
    \centering
    \includegraphics[width=1.\linewidth]{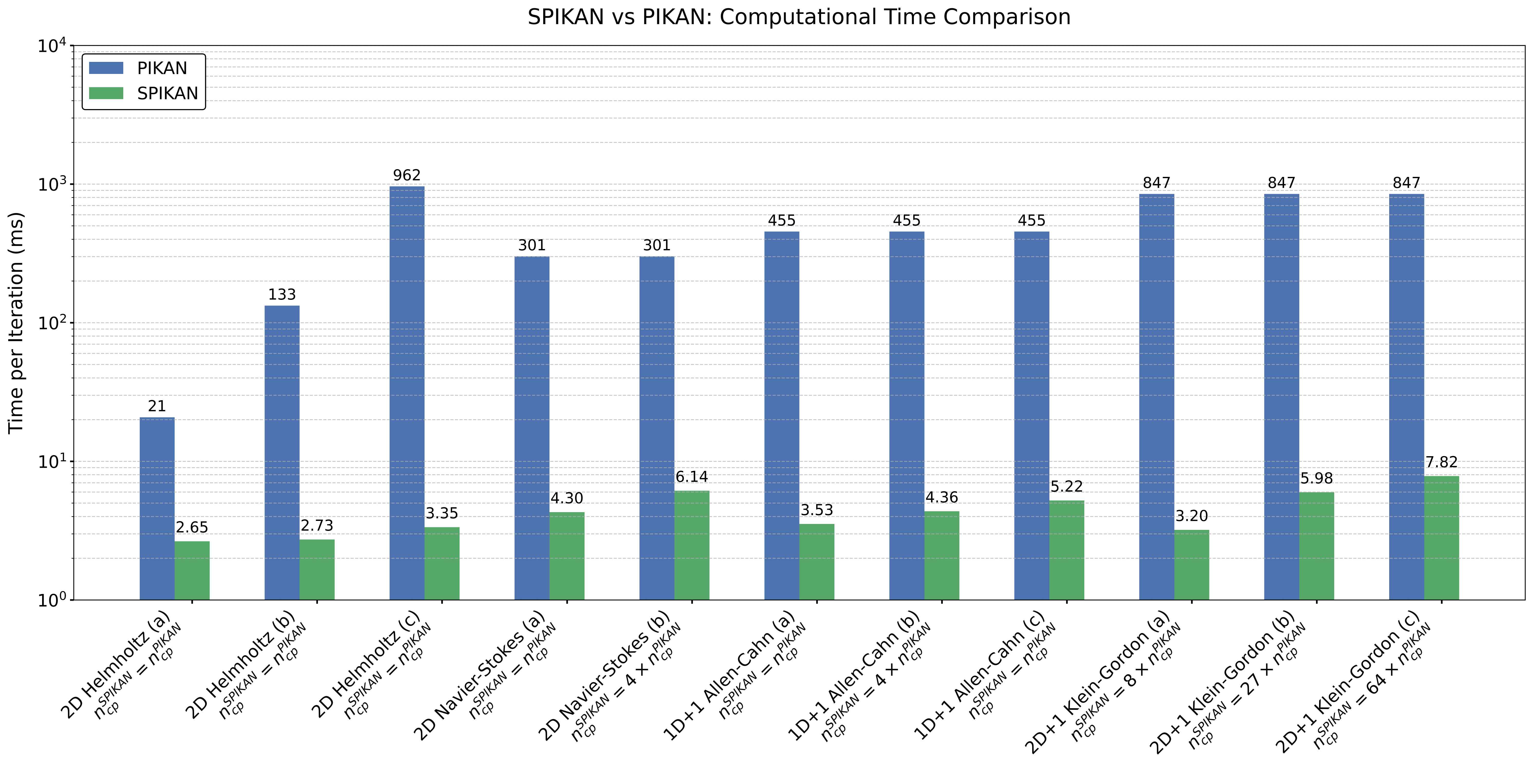}
    \caption{Comparison of computational time for all tests performed in Sec.\ref{sec:results}. A detailed description of each case is shown in Tables \ref{tab:2dhelmholtz}-\ref{tab:klein_gordon_results}.}
    \label{fig:computational_time_comparison}
\end{figure}

We summarize the computational times of all results in Fig.~\ref{fig:computational_time_comparison}. In all tested cases, SPIKANs achieved speedups ranging from $8\times$ to $287\times$, while maintaining $L_2$ error comparable to or better than that obtained with PIKANs. This result demonstrates that even for small numbers of collocation points and network sizes, leveraging SPIKANs remains advantageous.

An important aspect of SPIKANs involves the need for a factorizable grid of collocation points. While this requirement can be limiting for certain applications, we believe that the use of factorizable grids along with immersed boundary forcing terms, as recently proposed in \cite{sundar2024physics}, can help alleviate this issue. Furthermore, by using partition of unity functions to train an ensemble of KANs, as introduced in \cite{howard2024finite}, one could apply different factorizable grids in various areas of the domain. An important feature of SPIKANs is that they can be applied in addition to techniques for improving training and performance for KANs, including the use of other basis functions \cite{bozorgasl2024wav, so2024higher}, adaptive weighting such as residual-based attention \cite{anagnostopoulos2024residual, shukla2024comprehensive}, and adaptive grid refinement \cite{rigas2024adaptive}.
In future work, we will further explore ways to improve the accuracy of this architecture by leveraging multi-fidelity training, as in \cite{howard2023stacked} and extensions to operator learning with DeepOKANs, along the lines of separable DeepONets \cite{mandl2024separable}. 

\section{Code and data availability}
All code, trained models, and data required to replicate the examples presented in this paper will be released upon publication.

\section{Acknowledgements}
This project was completed with support from the U.S. Department of Energy, Advanced Scientific Computing Research program, under the Uncertainty Quantification for Multifidelity Operator Learning (MOLUcQ) project (Project No. 81739). The computational work was performed using PNNL Institutional Computing at Pacific Northwest National Laboratory. Pacific Northwest National Laboratory (PNNL) is a multi-program national laboratory operated for the U.S. Department of Energy (DOE) by Battelle Memorial Institute under Contract No. DE-AC05-76RL01830.

\bibliographystyle{my-elsarticle-num}
\bibliography{refs.bib}

\begin{thebibliography}{10}
\expandafter\ifx\csname url\endcsname\relax
  \def\url#1{\texttt{#1}}\fi
\expandafter\ifx\csname urlprefix\endcsname\relax\def\urlprefix{URL }\fi
\expandafter\ifx\csname href\endcsname\relax
  \def\href#1#2{#2} \def\path#1{#1}\fi

\bibitem{raissi2019physics}
M.~Raissi, P.~Perdikaris, G.~E. Karniadakis, Physics-informed neural networks: A deep learning framework for solving forward and inverse problems involving nonlinear partial differential equations, \emph{Journal of Computational physics} 378 (2019) 686--707.

\bibitem{karniadakis2021physics}
G.~E. Karniadakis, I.~G. Kevrekidis, L.~Lu, P.~Perdikaris, S.~Wang, L.~Yang, Physics-informed machine learning, \emph{Nature Reviews Physics} 3~(6) (2021) 422--440.

\bibitem{liu2024kan}
Z.~Liu, Y.~Wang, S.~Vaidya, F.~Ruehle, J.~Halverson, M.~Solja{\v{c}}i{\'c}, T.~Y. Hou, M.~Tegmark, Kan: {Kolmogorov-Arnold} networks, \emph{arXiv preprint arXiv:2404.19756}.

\bibitem{liu2024kan20}
Z.~Liu, P.~Ma, Y.~Wang, W.~Matusik, M.~Tegmark, Kan 2.0: Kolmogorov-arnold networks meet science, \emph{arXiv preprint arXiv:2408.10205}.

\bibitem{vaca2024kolmogorov}
C.~J. Vaca-Rubio, L.~Blanco, R.~Pereira, M.~Caus, {Kolmogorov-Arnold} networks ({KANs}) for time series analysis, \emph{arXiv preprint arXiv:2405.08790}.

\bibitem{samadi2024smooth}
M.~E. Samadi, Y.~M{\"u}ller, A.~Schuppert, Smooth {Kolmogorov-Arnold} networks enabling structural knowledge representation, \emph{arXiv preprint arXiv:2405.11318}.

\bibitem{howard2024finite}
A.~A. Howard, B.~Jacob, S.~H. Murphy, A.~Heinlein, P.~Stinis, Finite basis {Kolmogorov-Arnold} networks: domain decomposition for data-driven and physics-informed problems, \emph{arXiv preprint arXiv:2406.19662}.

\bibitem{shukla2024comprehensive}
K.~Shukla, J.~D. Toscano, Z.~Wang, Z.~Zou, G.~E. Karniadakis, A comprehensive and fair comparison between mlp and kan representations for differential equations and operator networks, \emph{arXiv preprint arXiv:2406.02917}.

\bibitem{abueidda2024deepokan}
D.~W. Abueidda, P.~Pantidis, M.~E. Mobasher, Deepokan: Deep operator network based on kolmogorov arnold networks for mechanics problems, \emph{arXiv preprint arXiv:2405.19143}.

\bibitem{rigas2024adaptive}
S.~Rigas, M.~Papachristou, T.~Papadopoulos, F.~Anagnostopoulos, G.~Alexandridis, Adaptive training of grid-dependent physics-informed kolmogorov-arnold networks, \emph{arXiv preprint arXiv:2407.17611}.

\bibitem{ranasinghe2024ginn}
N.~Ranasinghe, Y.~Xia, S.~Seneviratne, S.~Halgamuge, Ginn-kan: Interpretability pipelining with applications in physics informed neural networks, \emph{arXiv preprint arXiv:2408.14780}.

\bibitem{wang2024kolmogorov}
Y.~Wang, J.~Sun, J.~Bai, C.~Anitescu, M.~S. Eshaghi, X.~Zhuang, T.~Rabczuk, Y.~Liu, Kolmogorov arnold informed neural network: A physics-informed deep learning framework for solving pdes based on kolmogorov arnold networks, \emph{arXiv preprint arXiv:2406.11045}.

\bibitem{toscano2024inferring}
J.~D. Toscano, T.~K{\"a}ufer, M.~Maxey, C.~Cierpka, G.~E. Karniadakis, Inferring turbulent velocity and temperature fields and their statistics from lagrangian velocity measurements using physics-informed kolmogorov-arnold networks, \emph{arXiv preprint arXiv:2407.15727}.

\bibitem{patra2024physics}
S.~Patra, S.~Panda, B.~K. Parida, M.~Arya, K.~Jacobs, D.~I. Bondar, A.~Sen, Physics informed kolmogorov-arnold neural networks for dynamical analysis via efficent-kan and wav-kan, \emph{arXiv preprint arXiv:2407.18373}.

\bibitem{shuai2024physics}
H.~Shuai, F.~Li, Physics-informed kolmogorov-arnold networks for power system dynamics, \emph{arXiv preprint arXiv:2408.06650}.

\bibitem{so2024higher}
C.~C. So, S.~P. Yung, Higher-order-relu-kans (hrkans) for solving physics-informed neural networks (pinns) more accurately, robustly and faster, \emph{arXiv preprint arXiv:2409.14248}.

\bibitem{bozorgasl2024wav}
Z.~Bozorgasl, H.~Chen, Wav-kan: Wavelet {Kolmogorov-Arnold} networks, \emph{arXiv preprint arXiv:2405.12832}.

\bibitem{li2024kolmogorov}
Z.~Li, {Kolmogorov-Arnold} networks are radial basis function networks, \emph{arXiv preprint arXiv:2405.06721}.

\bibitem{ss2024chebyshev}
S.~SS, Chebyshev polynomial-based {Kolmogorov-Arnold} networks: An efficient architecture for nonlinear function approximation, \emph{arXiv preprint arXiv:2405.07200}.

\bibitem{yu2024sinc}
T.~Yu, J.~Qiu, J.~Yang, I.~Oseledets, Sinc {Kolmogorov-Arnold} network and its applications on physics-informed neural networks, \emph{arXiv preprint arXiv:2410.04096}.

\bibitem{cho2024separable}
J.~Cho, S.~Nam, H.~Yang, S.-B. Yun, Y.~Hong, E.~Park, Separable physics-informed neural networks, \emph{Advances in Neural Information Processing Systems} 36.

\bibitem{mcclenny2023self}
L.~D. McClenny, U.~M. Braga-Neto, Self-adaptive physics-informed neural networks, \emph{Journal of Computational Physics} 474 (2023) 111722.

\bibitem{anagnostopoulos2024residual}
S.~J. Anagnostopoulos, J.~D. Toscano, N.~Stergiopulos, G.~E. Karniadakis, Residual-based attention in physics-informed neural networks, \emph{Computer Methods in Applied Mechanics and Engineering} 421 (2024) 116805.

\bibitem{chen2024self}
W.~Chen, A.~A. Howard, P.~Stinis, Self-adaptive weights based on balanced residual decay rate for physics-informed neural networks and deep operator networks, \emph{arXiv preprint arXiv:2407.01613}.

\bibitem{jax2018github}
J.~Bradbury, R.~Frostig, P.~Hawkins, M.~J. Johnson, C.~Leary, D.~Maclaurin, G.~Necula, A.~Paszke, J.~Vander{P}las, S.~Wanderman-{M}ilne, Q.~Zhang, \href{http://github.com/google/jax}{{JAX}: composable transformations of {P}ython+{N}um{P}y programs} (2018).
\newline\urlprefix\url{http://github.com/google/jax}

\bibitem{Rigas_jaxKAN_A_JAX-based_2024}
S.~Rigas, M.~Papachristou, \href{https://github.com/srigas/jaxKAN}{{jaxKAN: A JAX-based implementation of {Kolmogorov-Arnold} networks}} (May 2024).
\newline\urlprefix\url{https://github.com/srigas/jaxKAN}

\bibitem{patankar1983calculation}
S.~V. Patankar, D.~B. Spalding, A calculation procedure for heat, mass and momentum transfer in three-dimensional parabolic flows, in: Numerical prediction of flow, heat transfer, turbulence and combustion, Elsevier, 1983, pp. 54--73.

\bibitem{ghia1982high}
U.~Ghia, K.~N. Ghia, C.~Shin, High-re solutions for incompressible flow using the {Navier-Stokes} equations and a multigrid method, \emph{Journal of computational physics} 48~(3) (1982) 387--411.

\bibitem{wight2020solving}
C.~L. Wight, J.~Zhao, Solving {Allen-Cahn} and {Cahn-Hilliard} equations using the adaptive physics informed neural networks, \emph{arXiv preprint arXiv:2007.04542}.

\bibitem{mattey2022novel}
R.~Mattey, S.~Ghosh, A novel sequential method to train physics informed neural networks for {Allen-Cahn} and {Cahn-Hilliard} equations, \emph{Computer Methods in Applied Mechanics and Engineering} 390 (2022) 114474.

\bibitem{rohrhofer2022role}
F.~M. Rohrhofer, S.~Posch, C.~G{\"o}{\ss}nitzer, B.~C. Geiger, On the role of fixed points of dynamical systems in training physics-informed neural networks, \emph{arXiv preprint arXiv:2203.13648}.

\bibitem{howard2024multifidelity}
A.~Howard, Y.~Fu, P.~Stinis, A multifidelity approach to continual learning for physical systems, \emph{Machine Learning: Science and Technology} 5~(2) (2024) 025042.

\bibitem{heinlein2024multifidelity}
A.~Heinlein, A.~A. Howard, D.~Beecroft, P.~Stinis, Multifidelity domain decomposition-based physics-informed neural networks for time-dependent problems, \emph{arXiv preprint arXiv:2401.07888}.

\bibitem{wang2022respecting}
S.~Wang, S.~Sankaran, P.~Perdikaris, Respecting causality is all you need for training physics-informed neural networks, \emph{arXiv preprint arXiv:2203.07404}.

\bibitem{sundar2024physics}
R.~Sundar, D.~Majumdar, D.~Lucor, S.~Sarkar, Physics-informed neural networks modelling for systems with moving immersed boundaries: Application to an unsteady flow past a plunging foil, \emph{Journal of Fluids and Structures} 125 (2024) 104066.

\bibitem{howard2023stacked}
A.~A. Howard, S.~H. Murphy, S.~E. Ahmed, P.~Stinis, Stacked networks improve physics-informed training: applications to neural networks and deep operator networks, \emph{arXiv preprint arXiv:2311.06483}.

\bibitem{mandl2024separable}
L.~Mandl, S.~Goswami, L.~Lambers, T.~Ricken, Separable deeponet: Breaking the curse of dimensionality in physics-informed machine learning, \emph{arXiv preprint arXiv:2407.15887}.

\end{thebibliography}

\end{document}